\documentclass[sigconf,natbib,screen]{acmart}

\usepackage{graphicx}
\usepackage{url}
\usepackage{xspace}
\usepackage{latexsym}
\usepackage{paralist}
\usepackage{amsmath}
\usepackage{amssymb}
\usepackage{multirow}
\usepackage{subfigure}
\usepackage{booktabs}
\usepackage{color}
\usepackage{amsfonts}
\usepackage{threeparttable}
\usepackage{xAlgorithm}
\usepackage[skip=0pt]{caption}
\usepackage{flushend}

\AtBeginDocument{%
  \providecommand\BibTeX{{%
    \normalfont B\kern-0.5em{\scshape i\kern-0.25em b}\kern-0.8em\TeX}}}

\newcommand{\goodgap}{\hspace{\subfigtopskip}\hspace{\subfigbottomskip}}

\newtheorem{Example}{\textbf{Example}}

\newcommand{\our}{\textsf{DAT}\xspace}
\newcommand{\rsn}{\textsf{RSNs}\xspace}
\newcommand{\mtranse}{\textsf{MTransE}\xspace}
\newcommand{\iptranse}{\textsf{IPTransE}\xspace}
\newcommand{\jape}{\textsf{JAPE}\xspace}
\newcommand{\bootea}{\textsf{BootEA}\xspace}
\newcommand{\gcn}{\textsf{GCN-Align}\xspace}

\newcommand{\gm}{\textsf{GM-Align}\xspace}
\newcommand{\mul}{\textsf{MultiKE}\xspace}
\newcommand{\mc}{\textsf{MuGNN}\xspace}

\newcommand{\rd}{\textsf{RDGCN}\xspace}
\newcommand{\hgcn}{\textsf{HGCN}\xspace}
\newcommand{\kecg}{\textsf{KECG}\xspace}
\newcommand{\te}{\textsf{TransEdge}\xspace}
\newcommand{\sota}{state-of-the-art\xspace}
\let\vec\mathbf

\begin{document}
	
\title[Degree-Aware Alignment for Entities in Tail]{Degree-Aware Alignment for Entities in Tail}

\author{Weixin Zeng}
\affiliation{%
	\institution{National University of Defense Technology}
}
\email{zengweixin13@nudt.edu.cn}

\author{Xiang Zhao}
\authornote{Corresponding author.}
\affiliation{%
	\institution{National University of Defense Technology}
}
\email{xiangzhao@nudt.edu.cn}

\author{Wei Wang}
\affiliation{%
	\institution{The University of New South Wales}
}
\email{weiw@cse.unsw.edu.au}

\author{Jiuyang Tang}
\affiliation{%
	\institution{National University of Defense Technology}
}
\email{jiuyang\_tang@nudt.edu.cn}

\author{Zhen Tan}
\affiliation{%
	\institution{National University of Defense Technology}
}
\email{tanzhen08a@nudt.edu.cn}
\renewcommand{\shortauthors}{Zeng et al.}
\fancyhead{}

\begin{abstract}
  Entity alignment (EA) is to discover equivalent entities in knowledge graphs (KGs), which bridges heterogeneous sources of information and facilitates the integration of knowledge. Existing EA solutions mainly rely on structural information to align entities, typically through KG embedding. Nonetheless, in real-life KGs, only a few entities are densely connected to others, and the rest majority possess rather sparse neighborhood structure. We refer to the latter as \emph{long-tail entities}, and observe that such phenomenon arguably limits the use of structural information for EA.
  
  To mitigate the issue, we revisit and investigate into the conventional EA pipeline in pursuit of elegant performance. For pre-alignment, we propose to amplify long-tail entities, which are of relatively weak structural information, with \emph{entity name} information that is generally available (but overlooked) in the form of concatenated power mean word embeddings. 
  For alignment, under a novel complementary framework of consolidating structural and name signals, we identify entity's \emph{degree} as important guidance to effectively fuse two different sources of information. To this end, a degree-aware \emph{co-attention} network is conceived, which dynamically adjusts the significance of features in a degree-aware manner.
  For post-alignment, we propose to complement original KGs with facts from their counterparts by using confident EA results as anchors via \emph{iterative} training. Comprehensive experimental evaluations validate the superiority of our proposed techniques.
\end{abstract}

\keywords{Entity alignment; Long-tail; Co-attention; Iterative training}

\maketitle

\section{Introduction}

Over recent years, a large number of Knowledge Graphs (KGs), e.g.,
\texttt{YAGO}~\cite{DBLP:conf/www/SuchanekKW07},
\texttt{DBpedia}~\cite{DBLP:conf/semweb/AuerBKLCI07}, \texttt{Knowledge
	Vault}~\cite{DBLP:conf/kdd/0001GHHLMSSZ14} and
\texttt{NELL}~\cite{DBLP:conf/aaai/CarlsonBKSHM10}, have been constructed, which
contribute significantly to the development of intelligent information
services.
In addition, there is a growing number of domain-specific KGs, such as medical
KG~\footnote{\url{https://flowhealth.com/}} and scientific
KG~\footnote{\url{https://www.aminer.cn/scikg}}. Nevertheless, a single KG can
never reach perfect coverage or being 100\% correct, due to the
inevitable trade-off that the KG construction process needs to take between \emph{coverage} and \emph{correctness}~\cite{DBLP:journals/semweb/Paulheim17}.

An effective way to automatically and efficiently increase the coverage and correctness of KGs
is by integrating knowledge from other KGs. This is because KGs constructed
independently and/or from independent sources generally provide redundancy
and complementary information, so that the integrated KG is expected to be better
in both coverage and correctness. 
For instance, a general KG constructed from
Web pages might only contain brand names of medicine, while more details of it (e.g., generic names of medicine)
can be found in a medical KG constructed from
medical literature. To incorporate knowledge from external
KGs into the original KG, the first and the most crucial step, is to align KGs.
As such, recent efforts have
been devoted to entity alignment (EA)~\cite{IJCAI19,ACL19,DBLP:conf/aaai/TrisedyaQZ19}, 
and these aligned entities serve as pivots to connect
KGs and lay foundation for downstream tasks.

Current solutions to EA mainly rely on graph
structure of KGs~\cite{DBLP:conf/www/PeiYHZ19,DBLP:conf/icml/GuoSH19,DBLP:conf/ijcai/ChenTYZ17,DBLP:conf/emnlp/WangLLZ18,DBLP:conf/ijcai/SunHZQ18},
which take for granted that equivalent entities possess similar neighborhood structures.
On some synthetic datasets extracted from large-scale KGs, these methods have achieved state-of-the-art
performance~\cite{DBLP:conf/ijcai/SunHZQ18,DBLP:conf/acl/CaoLLLLC19,DBLP:conf/ijcai/ZhuZ0TG19}.
However, recent study pointed out that those \emph{synthetic} datasets are much denser than KGs in real life, and existing EA methods are \emph{not} able to yield satisfactory results on datasets with real-life distributions~\cite{DBLP:conf/icml/GuoSH19}.

\begin{figure} [b]
	\centering
	\includegraphics[width=\linewidth]{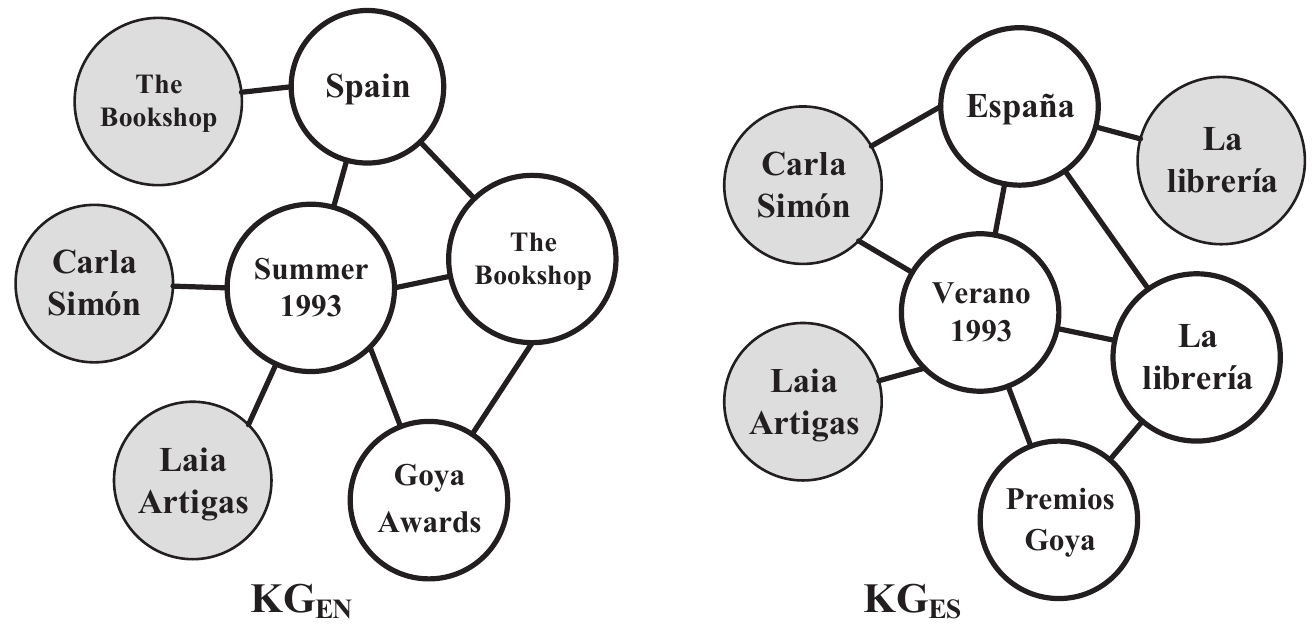}
	\caption{An example of EA. Nodes in grey (resp. white) are long-tail (resp. popular) entities (relation names and other entities are omitted in the interest
		of space).}\label{fig:1}
\end{figure}

In particular,~\citet{DBLP:conf/icml/GuoSH19} revealed that nearly half
of the entities in real-life KGs are connected to less than three other
entities. We refer to them as \emph{long-tail entities}, and these entities
in tail make the KG a relatively sparse graph. This follows our perception of real-life
KGs that only very few entities are frequently accessed, and possess detailed
attributes and rich connections, whereas the majority are left under-explored
and come with little structural information.
As a consequence, existing EA
methods that solely rely on structural information tend to take a toll on their
capability of accurate alignment, as demonstrated by Example~\ref{eg1}.

\begin{Example}\label{eg1}%
	In Figure~\ref{fig:1} is a partial English KG (KG$_{\text{EN}}$) and a
	partial Spanish KG (KG$_{\text{ES}}$) concerning the film
	\texttt{Summer 1993}. 
	Note that the entities \texttt{The Bookshop} and \texttt{La Librer\'{i}a} in grey
	describe the original novel, while those in white depict the film.
	
	During aligning entities of high degrees, e,g.,
	\texttt{Spain} and \texttt{Espa\~{n}a}, structural information is of great help; however, as to long-tail entities, e.g., \texttt{Carla Sim\'{o}n} in KG$_{\text{EN}}$,
	structural information may suggest \texttt{Laia Artigas} in KG$_{\text{ES}}$ as its match, since they have a single link to \texttt{Summer 1993} and \texttt{Verano 1993}, respectively.
\end{Example}

The example unveils the shortcoming of solely relying on structural
information for EA, which renders existing EA methods sup-optimal, even infeasible for long-tail entities.
Hence, we are motivated to revisit the key phases of EA pipeline, and address the challenge of EA when structure information is insufficient.

For \emph{pre-alignment} phase, we seek additional signals that can benefit EA, and discover a source of
information from entity names. It is generally available among real-life entities, yet has been overlooked by existing research.
For instance, for the long-tail entity \texttt{Carla Sim\'{o}n}
in KG$_{\text{EN}}$, introducing entity name information would easily help
locate the correct mapping---\texttt{Carla Sim\'{o}n} in KG$_{\text{ES}}$.
Thus, entity name information tends to serve as a \emph{complementing}
view to the commonly used structural information. 
In fact, name embeddings (in the form of averaged word embedding) have been incorporated before~\cite{ACL19,DBLP:conf/ijcai/WuLF0Y019,wu2019jointly}, which were plainly used to fill up the initial feature matrices for learning \emph{structural representation}; in contrast, we explicitly harness entity names as another channel of signal, parallel to structure information~\cite{IJCAI19}, by encoding via concatenated power mean word embeddings.

For \emph{alignment} phase, we explore to judiciously fuse the two aforementioned signals, based on the observation that for entities of different degrees, the importance of
structural and name information varies.
In Example~\ref{eg1}, when
aligning long-tail entity \texttt{Carla Sim\'{o}n} in KG$_{\text{EN}}$, its
limited neighboring structure is less useful than entity name information; distinctively, structure plays a more significant role for mapping popular
entities like the film \texttt{La Librer\'{i}a}, typically in the presence of
ambiguous entity names (i.e., both the film \texttt{La Librer\'{i}a} and
the novel \texttt{La Librer\'{i}a} share the same name). In general, a plausible intuition
is that \emph{the lower (resp. higher) the degree}, \emph{the more important the signal from entity name (resp. neighboring structure)}. To precisely
capture the non-linear dynamics between these two signals, we devise a co-attention network to determine the weights of different signals under the guidance of entity degrees.
Noteworthily,~\citet{DBLP:conf/www/PeiYHZ19} introduced degrees with the intention of correcting the undesirable bias from \emph{structural embedding} methods that place entities with similar degrees closely; our motivation is different in that degrees are leveraged to compute pair-wise similarities (rather than individual embeddings).

For \emph{post-alignment} phase, we propose to substantially enhance structural information of KGs by looking at and referencing each other recurrently.
Long-tail entities fall short of structural information in their
original KG (source KG), but the KG to be aligned (target KG) may possess such information
complementarily.
For example, for the entity \texttt{Carla Sim\'{o}n}, the fact that \texttt{Carla Sim\'{o}n} is from \texttt{Espa\~{n}a} is missing in KG$_{\text{EN}}$, which is yet seen in KG$_{\text{ES}}$. If the source KG can harvest it from the counterpart of the target KG (after pairing the surrounding entities), alignment could be ameliorated.
Inspired by the positive effect of initial completion of KGs using rules~\cite{DBLP:conf/acl/CaoLLLLC19}, we conceive an \emph{iterative} training procedure with KG completion embedded, which takes confident EA results in each round as anchors and replenishes possible missing relations to enhance the current KGs. As a net effect, 
these KGs are enriched, from which better structural embeddings could be learned, and the matching signal could propagate to long-tail entities---previously difficult in a single shot but now may become easier to align.

\subsubsection*{Contribution}
In short, our contribution can be summarized as
\begin{itemize}
	\item We identify the deficiency of existing EA methods in aligning long-tail
	entities, largely due to sole reliance on structure. We approach the limit by
	(1) introducing a complementary signal from entity names in the form of concatenated power mean word embeddings; and
	(2) conceiving an effective way via degree-aware co-attention mechanism to dynamically fuse name and structural signals. 
	
	\item We propose to reduce long-tail entities through augmenting relational structure via KG completion embedded into an iterative self-training strategy, which is realized by taking confident EA results as anchors and using each other KG as references. The strategy enhances the performance of EA and the coverage of KGs simultaneously.
	
	\item The techniques constitute a novel framework, namely, \our. We empirically evaluate the implementation of \our on both mono-lingual and cross-lingual EA
	tasks against state-of-the-art methods, and the comparative results and
	ablation analysis demonstrate the superiority of \our.
\end{itemize}

\subsubsection*{Organization}
Section~\ref{ref} overviews related work. In Section~\ref{ana}, we analyse the
long-tail phenomenon in EA. \our and its components are elaborated in
Section~\ref{method}. Section~\ref{exp} introduces experimental settings,
evaluation results and detailed analysis, followed by conclusion in
Section~\ref{conclude}.

\begin{table*}
	\centering \small
	\caption{Degree distribution of entities in test set (the first KG in each KG pair) and results of \rsn}\label{tab:degreeinfo}
	\begin{tabular}{lcccccccccccc}
		\toprule
		\multirow{2}[4]{*}{Degree} & \multicolumn{3}{c}{EN-FR} & \multicolumn{3}{c}{EN-DE} & \multicolumn{3}{c}{DBP-WD} & \multicolumn{3}{c}{DBP-YG} \\
		\cmidrule{2-13}          & \#Total & \#Correct & Accuracy & \#Total & \#Correct & Accuracy & \#Total & \#Correct & Accuracy & \#Total & \#Correct & Accuracy \\
		\midrule
		1     & 2,660  & 380   & 14.29\% & 1,978  & 453   & 22.90\% & 1,341  & 276   & 20.58\% & 3,327  & 538   & 16.17\% \\
		2     & 2,540  & 699   & 27.52\% & 2,504  & 1,005  & 40.14\% & 2,979  & 800   & 26.85\% & 2,187  & 688   & 31.46\% \\
		3     & 1,130  & 408   & 36.11\% & 1,514  & 820   & 54.16\% & 1,789  & 600   & 33.54\% & 1,143  & 563   & 49.26\% \\
		>=4   & 3,120  & 1,803  & 57.79\% & 3,454  & 2,416  & 69.95\% & 3,341  & 2,093  & 62.65\% & 2,793  & 2,008  & 71.89\% \\
		\midrule
		All   & 9,450  & 3,290  & 34.81\% & 9,450  & 4,694  & 49.67\% & 9,450  & 3,769  & 39.88\% & 9,450  & 3,797  & 40.18\% \\
		\bottomrule
	\end{tabular}%
\end{table*}%

\begin{figure} [b]
	\centering
	\includegraphics[width=\linewidth]{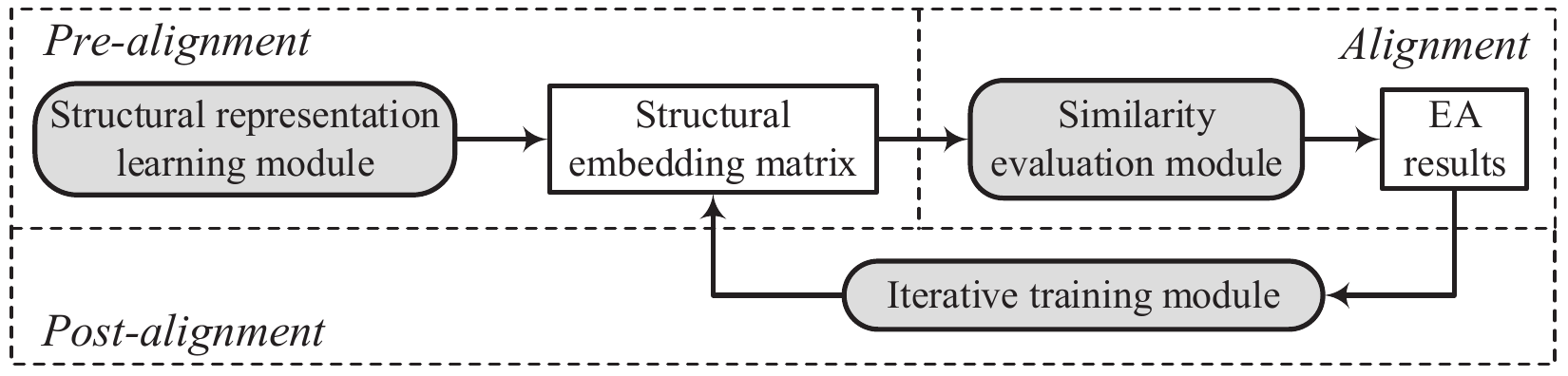}
	\caption{Conventional framework of EA}~\label{fig:old-framework}
\end{figure}

\section{Related Work}\label{ref}

\subsubsection*{Conventional EA framework}
The contributions of \sota methods can be described from the angle of a phased pipeline for EA.
For pre-alignment, one utilizes KG representation methods, e.g., \textsf{TransE}~\cite{DBLP:conf/ijcai/ChenTYZ17,DBLP:conf/ijcai/ZhuXLS17,DBLP:conf/ijcai/ChenTCSZ18} and \textsf{GCN}~\cite{DBLP:conf/emnlp/WangLLZ18}, to encode structural information and embed KGs into individual low-dimensional spaces.
Then for alignment, under the supervision of seed entity pairs, alignment results can be derived by evaluating and comparing the embedding spaces. 
Some methods~\cite{DBLP:conf/semweb/SunHL17,DBLP:conf/ijcai/SunHZQ18,DBLP:conf/icml/GuoSH19} directly project entities in different KGs into the same embedding space by fusing the training corpus in pre-alignment; then for alignment, according to some distance in the unified embedding space, equivalence in different KGs can also be found. 
For post-alignment, iterative strategies~\cite{DBLP:conf/ijcai/SunHZQ18,DBLP:conf/ijcai/ZhuXLS17} are employed to augment supervision signals by exploiting the results of alignment phase. In this way, structural embeddings are updated, and alignment can be recursively executed until reaching some stopping criterion. 
These techniques can be roughly summarized into a framework, depicted by Figure~\ref{fig:old-framework}.

\subsubsection*{Recent advancement on EA}
There are some very recent efforts aiming to overcome structural heterogeneity by devising more advanced structure learning models, e.g., topic graph matching~\cite{ACL19} and multi-channel graph neural network~\cite{DBLP:conf/acl/CaoLLLLC19}.
Particularly,~\citet{DBLP:conf/www/PeiYHZ19} improved the structural embedding with awareness of the degree difference by performing adversarial training.
Albeit, it still relies on structure for aligning signals, which tends to be ineffective when entities in both KGs are in tail; and moreover, degree information here is used to better learn structural embeddings, whereas our use of degrees is to better fuse two different and useful aligning signals---structural and name information.

Iterative strategies are beneficial to improving EA, but can be either time-consuming and biased towards one KG~\cite{DBLP:conf/ijcai/SunHZQ18}, or susceptible to introducing many false positive instances~\cite{DBLP:conf/ijcai/ZhuXLS17}, which can barely satisfy the real-life requirement. To strike a balance between precision and computational overhead, we implement the idea of iterative training with the inclusion of a KG completion module, such that structure is actually updated in each round around confident anchoring entity pairs. The strategy is of light weight and keeps the possible inclusion of incorrect pairs at a small number.

It can be seen that almost all the aforementioned embeddings learn from structural information, which can be sometimes insufficient, especially for long-tail entities.
In this connection, some proposed to incorporate \emph{attributes}~\cite{DBLP:conf/semweb/SunHL17,DBLP:conf/emnlp/WangLLZ18,DBLP:conf/aaai/TrisedyaQZ19,IJCAI19} to potentially compensate the shortage. 
Nevertheless, between 69\% and 99\% of instances in popular KGs lack at least one attribute that other entities in the same class have~\cite{DBLP:conf/wsdm/GalarragaRAS17}.
Similarly, \emph{entity descriptions}~\cite{DBLP:conf/ijcai/ChenTCSZ18} could be utilized to provide extra information, which is nonetheless often missing in many KGs.
In short, these efforts possibly strengthen the overall EA performance but tend to fall short in aligning entities in tail.
Entity name has also been tried, either as initial features for learning \emph{structural representation}~\cite{ACL19,DBLP:conf/ijcai/WuLF0Y019,wu2019jointly}, or together with other information for representation learning~\cite{IJCAI19}. 
In contrast, we consolidate the features on top of separate similarity matrices learned from structure and name information; empirical evaluation of different strategies will be described in Section~\ref{sec:results}.

\section{Impact of Long-tail Phenomenon}\label{ana}

\begin{figure*}
	\centering
	\includegraphics[width=0.72\linewidth]{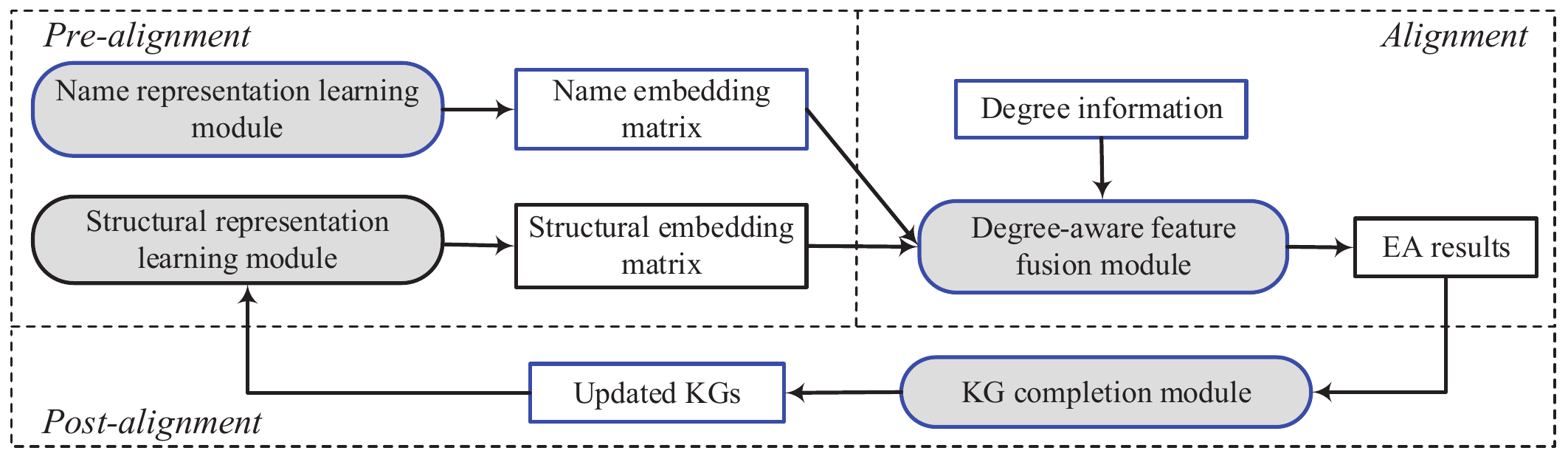}
	\caption{The framework of DAT}~\label{fig:framework}%
\end{figure*}

\subsubsection*{Task definition}

Given a source KG $G_1 = (E_1, R_1, T_1)$ and a target KG $G_2 = (E_2, R_2, T_2)$, where $E_1$ (resp. $E_2$)
represents source (resp. target) entities, $R$ denotes relations, $T \subseteq E\times R\times E$
represents triples. Denote the seed entity pairs as
$S = \{(e^{i}_1,e^{i}_{2})|e^{i}_1 = e^{i}_2, e^{i}_1\in E_1, e^{i}_{2}\in E_2\}$, $i \in [1, |S|]$, where $|\cdot|$ denote the cardinality of a set. EA
task is to find new EA pairs based on $S$ and
return the eventual results
$S^{'} = \{(e^{i}_1,e^{i}_{2})|e^{i}_1 = e^{i}_2, e^{i}_1\in E_1, e^{i}_{2}\in
E_2\}$, $i \in [1, \min\{|E_1|, |E_2|\}]$, where $=$ expresses that two entities are the same physical one.

A very recent work~\cite{DBLP:conf/icml/GuoSH19} points out that KGs in previous
EA datasets are too dense and the degree distributions deviate from real-life
KGs. Therefore, they establish a new EA benchmark that follows real-life
distribution. The evaluation benchmark consists of both cross-lingual datasets,
i.e., EN-FR, EN-DE, and mono-lingual datasets, i.e., DBP-WD, and DBP-YG. We show the
degree distributions of entities in test sets in Table~\ref{tab:degreeinfo}. The
degree of an entity is defined as the number of relational triples in which the
entity participates. We also report the amount of correctly aligned entities in
different degrees yielded by \rsn, the best solution in~\cite{DBLP:conf/icml/GuoSH19}.

It reads from Table~\ref{tab:degreeinfo} that in EN-FR and DBP-YG, entities
with degree less than three account for over 50\%, and nearly half of the
entities in EN-DE and DBP-WD are with degree 1 or 2. It confirms that the majority
of the entities in KG are long-tail and have very few connections to others. 
Moreover, it follows that the performance of long-tail entities is much
worse than those of higher degrees (the accuracies of high-degree entities
triple and even quadruple those of entities with degree 1), despite that \rsn is
the leading method on the benchmark.
This suggests that current methods fail to
effectively handle entities in tail, and hence, restrains the overall performance. Thereby, it is of significance to revisit the EA pipeline with a particular emphasis on long-tail entities. 

\section{Methodology}\label{method}

To provide an overview, we summarize the main components of the framework \our (\underline{d}egree-aware entity \underline{a}lignment in \underline{t}ail) in Figure~\ref{fig:framework}, where purple blue highlights the new designs in this research.
In pre-alignment, {\itshape Structural representation learning module} and {\itshape Name representation learning module} learn useful features of entities, i.e., name representation and structural representation; in alignment, these features are forwarded to {\itshape Degree-aware fusion module} for effective fusion and alignment under the guide of degree information.
In post-alignment, {\itshape KG completion module} aims to complete KGs with confident EA pairs in the results, and the augmented KGs are then again utilized in the next round iteratively.

Since {\itshape Structural representation learning module} has been extensively studied, we adopt the \sota model \rsn~\cite{DBLP:conf/icml/GuoSH19} for this purpose.
Given a structural embedding matrix $\vec Z \in \mathbb{R}^{n\times d_s}$, two
entities $e_1 \in G_1$ and $e_2 \in G_2$, their structural similarity $Sim_s(e_1,e_2)$ is the cosine similarity between $\vec {Z}(e_1)$ and $\vec Z(e_2)$, 
where $n$ denotes the number of all
entities in two KGs, $d_s$ is the dimension of structural embeddings, and
$\vec Z(e)$ denotes the embedding vector for entity $e$ (i.e.,
$\vec Z(e) = \mathbf{Z} \mathbf{e}$, where $\mathbf{e}$ is the one-hot encoding
of entity $e$). From the perspective of structure, the target entity with the
highest similarity to a source entity is returned as its alignment
result.

\subsection{Name representation learning}
Recall that structural information is of limited use for aligning long-tail entities. Distinct from existing efforts that try to exploit structures, we take another angle by seeking some signal that is beneficial and generally available to long-tail entities. 

To this end, we propose to incorporate the textual names of entities, a signal that has been largely overlooked by current embedding-based EA methods. 
In particular, the choice is alluring due to at least the following considerations:
\begin{inparaenum} [(1)]
	\item entity name can normally identify an entity, and given two entities, comparing their names may be the most intuitive way to judge the equivalence preliminarily; 
	\item most real-life entities possess a name, the ratio of which is much higher than that of other textual information (i.e., descriptions and attributes), which tends to be lacking for long-tail entities.
\end{inparaenum}

Despite that there are many classic approaches for measuring the
\emph{string similarity} between entity names, we go for \emph{semantic
	similarity} since it can still work when the vocabularies of KGs differ,
especially for the \emph{cross-lingual} scenario.
Specifically, we choose a general form of power mean embeddings~\cite{polya1952inequalities}, which encompasses many well-known means
such as the arithmetic mean, the geometric mean and the harmonic mean. Given a sequence of word embeddings,
$\vec w_1, \ldots, \vec w_l \in \mathbb{R}^d$, the power mean operation is formalized as 
\begin{equation}\label{pm}
(\frac{w_{1i}^p + \cdots + w_{li}^p}{l})^{1/ p}, \quad \forall i = 1, \ldots ,d, \quad p \in \mathbb{R}\cup{\pm\infty},
\end{equation}
where $l$ is the number of words and $d$ denotes the dimension of embeddings. It can be seen that setting $p$ to 1 results in the arithmetic mean, to 0 the geometric mean, to -1 the harmonic mean, to $+\infty$ the maximum operation and to $-\infty$ the minimum operation~\cite{DBLP:journals/corr/abs-1803-01400}.

Given a word embedding space $\mathbb{E}^i$, the embeddings of the words in the name of entity $s$ can be represented as $\vec W^i = [\vec w_1^i, \ldots, \vec w_l^i]\in \mathbb{R}^{l\times d^i}$. 
Correspondingly, $H_p(\vec W^i) \in \mathbb{R}^{d^i}$ denotes the power mean embedding vector after feeding $\vec w_1^i, \ldots, \vec w_l^i$ to Equation~(\ref{pm}).
To obtain summary statistics of entity $s$, we compute $K$ power means of $s$ and concatenate them to get the entity name representation $\vec s^i \in \mathbb{R}^{d^i\cdot K}$, i.e.,
\begin{equation}
\vec s^i = H_{p_1}(\vec W^i)\oplus\cdots\oplus H_{p_K}(\vec W^i),
\end{equation}
where $\oplus$ represents concatenation along rows, and $p_1, \ldots, p_K$ are $K$ different power mean values~\cite{DBLP:journals/corr/abs-1803-01400}. 

To get further representational power from different word embeddings, we generate the final entity name representation $\vec n_s$ by concatenating $\vec s^i$ obtained from different embedding spaces $ \mathbb{E}^i$:
\begin{equation}
\label{spacecat}
\vec n_s = \bigoplus\limits_{i} \vec s^i.
\end{equation}

Note that the dimensionality of this representation is $d_n = \sum_i d^i\cdot K$. 
The name embeddings of all entities can be denoted in matrix form as $\vec N \in \mathbb{R}^{n\times d_n}$.

Like word embeddings, semantically similar entity names will be placed adjacently in the entity name representation space. From the perspective of textual name, given two entities $e_1 \in G_1$ and $e_2 \in G_2$, their similarity $Sim_t(e_1,e_2)$ is the cosine similarity between $\vec N(e_1)$ and $\vec N(e_2)$, and the target entity with the highest similarity to a source entity is returned as its alignment result.

\subsubsection*{Discussion}
The concatenated power mean word embedding~\cite{DBLP:journals/corr/abs-1803-01400} is a better solution for representing entity name than averaged word embedding in that it can better summarize the useful signals in entity name~\footnote{For possible out-of-vocabulary (OOV) words, we skip them and use the embeddings of the rest to produce entity name embeddings.}.
The average of word embeddings discards large amount of information since variant names can be represented by similar averaged embedding. In contrast, the concatenation of different power means yields a more precise summary by reducing uncertainty about the semantic variation within a sentence.
This is further verified through empirical results in Section~\ref{abla}.

Note that for cross-lingual EA, we utilize pre-trained multilingual word
embeddings~\cite{DBLP:journals/corr/abs-1710-04087} that have already aligned
words from different languages into the same embedding space. This ensures that
entity names from various language sources also exist in the same semantic
space, which avoids designing an additional mapping function for aligning
multilingual embedding spaces.

Among others, the aforementioned method can be extended to support other textual information like attributes, without loss of generality.
An immediate solution is to concatenate attributes and entity name to form a ``sentence'' that describes the entity, which is then encoded by concatenated power mean word embeddings.
Nevertheless, the extension to incorporate additional information and more advanced adaptations are beyond the scope of this paper.

\subsection{Degree-aware co-attention feature fusion}
\begin{figure}
	\centering
	\includegraphics[width=1.05\linewidth]{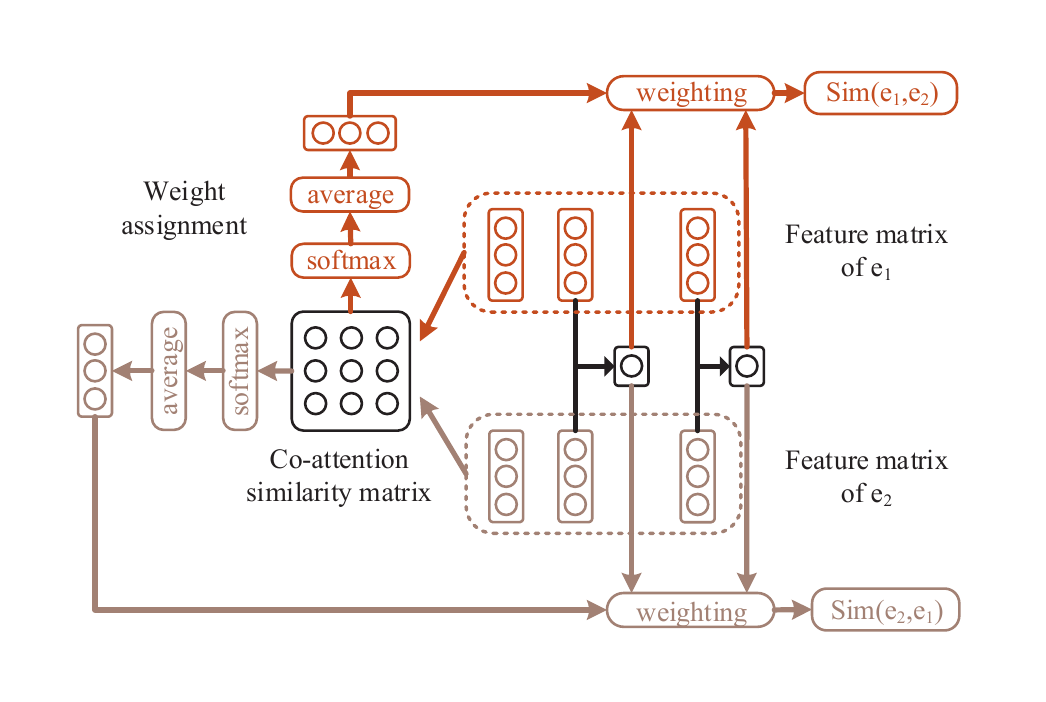}
	\caption{Degree-aware co-attention feature fusion}\label{fig:att}
\end{figure}

Different types of features characterize entity identities from different aspects. Thereby, a feature fusion module is of great significance to the effective combination of various signals.
Some propose to combine different embeddings into a unified representation space~\cite{IJCAI19}, whereas it requires additional training for unifying irrelevant features.
A more preferable strategy would be computing the similarity matrix within each feature-specific space first, and then combine the feature-specific similarity scores~\cite{DBLP:conf/emnlp/WangLLZ18,DBLP:conf/esws/PangZTT019}.
Nevertheless, for entities with different degrees, the contributions of different features vary.
Regarding long-tail entities that possess little structural information, entity name representation should be credited more; conversely, structural representation of popular entities is relatively more useful than entity name information.
To capture this dynamic change, inspired by the bi-attention mechanism~\cite{seo2016bidirectional}, we devise a degree-aware co-attention network, shown in Figure~\ref{fig:att}.

Formally, given the structural embedding matrix $\vec Z$, name embedding matrix $\vec N$, for every entity pair, $(e_1, e_2), e_1 \in G_1, e_2 \in G_2$, the similarity score between $e_1$ and $e_2$ is calculated, which is then utilized for determining the alignment result.
In order to compute the overall similarity, we first calculate the feature-specific similarity scores between $e_1$ and $e_2$, i.e., $Sim_s(e_1,e_2)$ and $Sim_t(e_1,e_2)$, as introduced in previous subsections. Then our degree-aware co-attention network aims to determine the weights for $Sim_s(e_1,e_2)$ and $Sim_t(e_1,e_2)$ by integrating degree information. This network comprises three stages---feature matrix construction, co-attention similarity matrix calculation, and weight assignment.

\subsubsection*{Feature matrix construction}
In addition to entity name and structural information, we incorporate entity degree information to construct feature matrix for each entity.
In specific, entity degrees are represented as one-hot vectors of all possible degree values, and then forwarded to a fully-connected layer to generate the continuous degree vector.
For instance, the degree vector of $e_1$ is $\vec g_{e_1} = \vec M \cdot \vec h_{e_1} \in \mathbb{R}^{d_g}$, where $\vec h_{e_1}$ is its one-hot representation, $\vec M$ is the weight matrix in the fully-connected layer and $d_g$ represents the dimension of degree vector.
This continuous degree vector, along with structural and entity name representations, is stacked to form an entity's feature matrix. For entity $e_1$,
\begin{equation}
\vec F_{e_1} = [\vec N(e_1); \vec Z(e_1); \vec g_{e_1}] \in \mathbb{R}^{3 \times d_m},
\end{equation}
where $;$ denotes the concatenation along \emph{columns}, $d_m = \max\{d_n, d_s, d_g\}$ and we pad the missing values with 0s.

\subsubsection*{Co-attention similarity matrix calculation}
To model the interaction between $\vec F_{e_1}$ and $\vec F_{e_2}$, as well as highlight important features, we build a co-attention matrix $\vec S \in \mathbb{R}^{3 \times 3}$, where the similarity between the $i$-th feature of $e_1$ and the $j$-th feature of $e_2$ is computed by
\begin{equation}
\vec S_{ij} = \alpha(\vec F_{e_1}^{i:}, \vec F_{e_2}^{j:}) \in \mathbb{R},
\end{equation}
where $\vec F_{e_1}^{i:}$ is the $i$-th row vector and $\vec F_{e_2}^{j:}$ is the $j$-th row vector, $i=1,2,3; j=1,2,3$.
$\alpha(\vec u, \vec v) = \vec w^{\top}(\vec u\oplus\vec v\oplus(\vec u\circ \vec v))$ is a trainable scalar function that encodes the similarity, where $\vec w \in \mathbb{R}^{3d_m}$ is a trainable weight vector, $\circ$ is the element-wise multiplication. Note that the implicit multiplication is a matrix multiplication.

\subsubsection*{Weight assignment}
The co-attention similarity matrix $\vec S$ is then utilized to obtain attention vectors, i.e., $\vec {att_1}$ and $\vec {att_2}$, in both directions.
$\vec {att_1}$ signifies which feature vectors of $e_1$ are most relevant to feature vectors of $e_2$, while $\vec {att_2}$ signifies which feature vectors of $e_2$ are most relevant to feature vectors of $e_1$. In specific, we feed $\vec S$ to a softmax layer. The resulting matrix is then compressed by an average layer to generate attention vectors. Note that column-wise operation in the softmax layer and row-wise operation in the average layer produce $\vec {att_1}$, while row-wise operation in the softmax layer and column-wise operation in the average layer produce $\vec {att_2}$.

Eventually, we multiply the feature-specific similarity scores with the attention values to obtain the final similarity score, 
\begin{equation}
Sim(e_1,e_2) = Sim_s(e_1,e_2)\cdot \vec {att_1}^s +  Sim_t(e_1,e_2)\cdot \vec {att_1}^t,
\end{equation}
where $\vec {att_1}^s$  and $\vec {att_1}^t$ are the corresponding weight values for structural and name similarity scores, respectively. Note that $Sim(e_1,e_2) \neq Sim(e_2,e_1)$ as they may have different attention weight vectors.

The co-attention feature fusion model is of low model complexity, with only parameters $\vec M$ and $\vec w$. In addition, it can also be easily adapted for including more features.

\subsubsection*{Training}
The training objective is to maximize the similarity scores of the training entity pairs, which can be converted to minimizing the following loss function:
\begin{equation}\label{eq:1} 
L = \sum_{(e_1,e_2)\in S} [\ -Sim(e_1,e_2) + \gamma]_+ + [\ -Sim(e_2,e_1) + \gamma]_+ ,
\end{equation}
where $[x]_+ = max\{0,x\}$, and $\gamma$ is a constant number. 

\subsubsection*{Discussion}
There could be other ways to implement degree-aware weighting, e.g., by using $\texttt{sigmoid}(\vec W \cdot [\vec N(e), \vec Z(e), \vec g_e])$, where $\vec W$ is the parameter.
Here we exploit a co-attention mechanism to consolidate different channels of signals with degree-aware weights, in order to demonstrate the effectiveness of leveraging degrees for EA in tail. 
More in-depth comparison with additional implementations 
is left for future work.

\subsection{Iterative KG completion}\label{method-ite}
Iterative self-training is an interesting idea to explore, which is shown to be effective~\cite{DBLP:conf/ijcai/SunHZQ18,DBLP:conf/ijcai/ZhuXLS17}.
However, existing research overlooked the opportunity to enrich structure information during the iterative process.
In fact, we found that although long-tail entities may lack structural information in the source KG, the target KG may possess such information in a complementary manner.
Intuitively, if we can replenish the original KG with facts from its counterpart by mining confident EA results as pseudo matching pairs anchoring the subgraphs, the KGs' structural sparsity could be mitigated. 
This can substantially improve the coverage of KGs and reduce the amount of long-tail entities. From the amplified KGs, the structural learning model can generate increasingly better structural embedding, which in turn leads to more accurate EA results in the subsequent rounds, and it is naturally iterative.

First, we detail the inclusion of EA pairs with high confidence, with an emphasis on avoiding false pairs that might adversely affect the model.
In particular, we devise a novel strategy for selecting EA pairs. 
For every given entity
$e_1 \in E_1 - S_1$ (in $G_1$ but not in the training set), suppose its most
similar entity in $G_2$ is $e_2$, its second most similar entity is $e_2^\prime$ and
the difference between the similarity scores is
$\Delta_1 \triangleq Sim(e_1, e_2)-Sim(e_1, e_2^\prime)$, if for $e_2$, its most similar entity in
$G_1$ is exactly $e_1$, its second most similar entity is $e_1^\prime$, the
difference between the similarity scores is
$\Delta_2 \triangleq Sim(e_2, e_1)-Sim(e_2, e_1^\prime)$, and $\Delta_1$, $\Delta_2$ are both above
a given threshold $\theta$, $(e_1, e_2)$ would be considered as a correct pair.
This is a relatively strong constraint, as it requires that (1) the similarity between the
two entities is the highest from both sides, respectively, and (2) there is a margin between the top-2 candidates.

After adding the highly
confident EA results to seed entity pairs, we use these entities 
($S_a$) to bridge two KGs and enrich the KGs with facts from the other side. For
instance, for a triple $t_1 \in
T_1$, if both its head and tail entities correspond to certain entries in
$S_a$, we replace entities in $t_1$ with the corresponding entities in
$E_2$ and add it to $T_2$. Although this is a very intuitive and simple
approach, it effectively reduces the number of long-tail entities and improves
the coverage of KGs in practice.
Finally, we utilize the augmented KGs for learning better structural
representation, which in turn contributes to the improvement of EA performance.
This iterative completion process lasts for $\zeta$ rounds.

\subsubsection*{Discussion}
Some EA methods also employ bootstrapping or iterative training strategies;
but they merely aim to enlarge the training signals to update the embeddings, without touching the actual structure of KGs. 
Besides, their procedures to select confident EA pairs are single-sided and
time-consuming~\cite{DBLP:conf/ijcai/SunHZQ18} or apt to introduce incorrect
instances~\cite{DBLP:conf/ijcai/ZhuXLS17}. 
In contrast, we substantially augment the KGs, and judiciously design the selection---pairing two entities only if they treat each other as the first priority, the superiority of which is witnessed in Section~\ref{feature-analysis}.

\section{Experiments}\label{exp}

This section reports the experiments with in-depth analysis~\footnote{The source code is available at \url{https://github.com/DexterZeng/DAT}}.
\subsection{Experiment setting}

\subsubsection*{Dataset}
We adopt $SRPRS$~\cite{DBLP:conf/icml/GuoSH19}, since the KG pairs thereof follow real-life degree distributions. 
This dataset was constructed by using inter-language links and reference links in \texttt{DBpedia}, and each entity has an equivalent entity in the other KG.
The detailed information is shown in Table~\ref{tab:data}. 30\% of entity pairs are harnessed for training.

\begin{table}
	\centering \small
	\caption{Statistics of $SRPRS$}\label{tab:data}%
	\begin{tabular}{lccc}
		\toprule
		Dataset & KGs & \#Triples & \#Entities \\
		\midrule
		\multirow{2}[2]{*}{EN-FR} & DBpedia (English) & 36,508 & 15,000 \\
		& DBpedia (French) & 33,532 & 15,000 \\
		\midrule
		\multirow{2}[1]{*}{EN-DE} & DBpedia (English) & 38,281 & 15,000 \\
		& DBpedia (German) & 37,069 & 15,000 \\
		\midrule
		\multirow{2}[0]{*}{DBP-WD} & DBpedia & 38,421 & 15,000 \\
		& Wikidata & 40,159 & 15,000 \\
		\midrule
		\multirow{2}[1]{*}{DBP-YG} & DBpedia & 33,571 & 15,000 \\
		& YAGO3 & 34,660 & 15,000 \\
		\bottomrule
	\end{tabular}%
\end{table}%

\subsubsection*{Parameter settings}
For the {\itshape Structural representation learning module}, we follow the settings in~\cite{DBLP:conf/icml/GuoSH19}, except for assigning $d_s$ to 300.
Regarding {\itshape Name representation learning module}, we set $\vec p = [p_1, \ldots, p_K]$ to $[1, \min, \max]$. 
For mono-lingual datasets, we merely use the \textsf{fastText} embeddings~\cite{bojanowski2017enriching} as the word embedding (i.e., only one embedding space in Equation~\ref{spacecat}). 
For cross-lingual datasets, the multilingual word embeddings are obtained from \textsf{MUSE}~\footnote{\url{https://github.com/facebookresearch/MUSE}}. 
Two word embedding spaces (from two languages) are used in Equation~\ref{spacecat}. 
As for {\itshape Degree-aware fusion module}, we set $d_g$ to 300, $\gamma$ to 0.8, batch size to 32. 
\textsf{Stochastic gradient descent} is harnessed to minimize the loss function, with learning rate set to 0.1, and we use early stopping to prevent over-fitting.
In {\itshape KG completion module}, $\theta$ is set to 0.05 and $\zeta$ is set to 3.

\begin{table*}[htbp]
	\begin{threeparttable}
		\centering \small
		\caption{Overall results of entity alignment}
		\begin{tabular}{lcccccccccccc}
			\toprule
			\multirow{2}[4]{*}{Methods} & \multicolumn{3}{c}{EN-FR} & \multicolumn{3}{c}{EN-DE} & \multicolumn{3}{c}{DBP-WD} & \multicolumn{3}{c}{DBP-YG} \\
			\cmidrule{2-13}          & Hits@1 & Hits@10 & MRR   & Hits@1 & Hits@10 & MRR   & Hits@1 & Hits@10 & MRR   & Hits@1 & Hits@10 & \multicolumn{1}{c}{MRR} \\
			\midrule
			\mtranse & 25.1  & 55.1  & 0.35  & 31.2  & 58.6  & 0.40   & 22.3  & 50.1  & 0.32  & 24.6  & 54.0    & 0.34 \\
			\iptranse & 25.5  & 55.7  & 0.36  & 31.3  & 59.2  & 0.41  & 23.1  & 51.7  & 0.33  & 22.7  & 50.0    & 0.32 \\
			\bootea & 31.3  & 62.9  & 0.42  & 44.2  & 70.1  & 0.53  & 32.3  & 63.1  & 0.42  & 31.3  & 62.5  & 0.42 \\
			\rsn  & 34.8  & 63.7  & 0.45 & 49.7  & 73.3  & 0.58 & 39.9  & 66.8  & 0.49 & 40.2  & 68.9  & 0.50 \\
			\mc & 13.1  & 34.2  & 0.20  & 24.5  & 43.1  & 0.31  & 15.1  & 36.6  & 0.22  & 17.5  & 38.1  & 0.24 \\
			\kecg & 29.8  & 61.6  & 0.40  & 44.4  & 70.7  & 0.54  & 32.3  & 64.6  & 0.43  & 35.0  & 65.1  & 0.45 \\
			\te &40.0  & 67.5  & 0.49  & 55.6  & 75.3  & 0.63  & 46.1  & 73.8  & 0.56  & 44.3  & 69.9  & 0.53 \\
			\midrule
			\gcn & 15.5  & 34.5  & 0.22  & 25.3  & 46.4  & 0.33  & 17.7  & 37.8  & 0.25  & 19.3  & 41.5  & 0.27 \\
			\jape  & 25.6  & 56.2  & 0.36  & 32.0    & 59.9  & 0.41  & 21.9  & 50.1  & 0.31  & 23.3  & 52.7  & 0.33 \\
			\rd & 67.5  & 76.9  & 0.71  & 78.3  & 88.4  & 0.82  & 83.4  & 90.7  & 0.86  & 85.8  & 93.8  & 0.89 \\
			\hgcn & 67.0  & 77.0 & 0.71 & 76.3  & 86.3  & 0.80  & 82.3  & 88.7  & 0.85 & 82.2  & 88.8  & 0.85 \\
			\gm$^1$ & 62.7  & -  & - & 67.7  & -  & - & 81.5  & -  & - & 82.8  & -  & - \\
			\our  & \textbf{75.8} & \textbf{89.9} & \textbf{0.81} & \textbf{87.6} & \textbf{95.5} & \textbf{0.90} & \textbf{92.6} & \textbf{97.7} & \textbf{0.94} & \textbf{94.0} & \textbf{98.5} & \textbf{0.96} \\
			\bottomrule
		\end{tabular}%
		\label{tab:EA}
		\begin{tablenotes}
			\footnotesize {
				\item[1] When running \gm, it is noted that entities without valid name embeddings are excluded from evaluation, and hence we consider that \gm fails to align these entities without specifying rankings, which leads to the lack of Hits@10 and MRR values.}
		\end{tablenotes}
	\end{threeparttable}
\end{table*}%

\begin{table*}
	\centering \small
	\caption{Hits@1 results by degrees}
	\begin{tabular}{lcccccccccccc}
		\toprule
		\multirow{2}[4]{*}{Degree} & \multicolumn{3}{c}{EN-FR} & \multicolumn{3}{c}{EN-DE} & \multicolumn{3}{c}{DBP-WD} & \multicolumn{3}{c}{DBP-YG} \\
		\cmidrule{2-13}          & \rsn    & \rd    & \our    & \rsn    & \rd    & \our    & \rsn    & \rd    & \our    & \rsn    & \rd    & \our \\
		\midrule
		1     & 14.3  & 56.5  & \textbf{57.4} & 22.9  & 75.8  & \textbf{83.1} & 20.6  & 80.3  & \textbf{86.7} & 16.2  & 84.2  & \textbf{89.5} \\
		2     & 27.5  & 64.4  & \textbf{72.4} & 40.1  & 78.6  & \textbf{84.3} & 26.9  & 84.0  & \textbf{90.8} & 31.5  & 80.9  & \textbf{92.6} \\
		3     & 36.1  & 77.3  & \textbf{82.9} & 54.2  & 77.4  & \textbf{88.4} & 33.5  & 75.2  & \textbf{88.2} & 49.3  & 89.2  & \textbf{97.0} \\
		$\geq$4   & 57.8  & 75.8  & \textbf{91.6} & 69.9  & 80.0  & \textbf{92.3} & 62.6  & 88.6  & \textbf{98.8} & 71.9  & 90.2  & \textbf{99.4} \\
		\midrule
		All   & 34.8  & 67.5  & \textbf{75.8} & 49.7  & 78.3  & \textbf{87.6} & 39.9  & 83.4  & \textbf{92.6}& 40.2  & 85.8  & \textbf{94.0} \\
		\bottomrule
	\end{tabular}%
	\label{tab:deg}%
\end{table*}%

\subsubsection*{Evaluation metric}
We utilize Hits@$k$ ($k$=1, 10) and mean reciprocal rank (MRR) as evaluation metrics.
For each source entity, entities in the other KG are ranked according to their similarity scores $Sim$ to the source entity in descending order.
Hits@$k$ reflects the percentage of correctly aligned entities in top-$k$ similar entities to source entities.
In particular, Hit@1 represents the accuracy of alignment results.
MRR, on the other hand, denotes the average of reciprocal ranks of ground truth results.
Note that higher Hits@$k$ and MRR indicate better performance. Unless otherwise specified, the results of Hits@$k$ are represented in percentages.
We denote the best performance in \textbf{bold} in the tables. 

\subsubsection*{Competitors}
Overall 13 state-of-the-art methods are involved in comparison.
The group that solely utilizes structural feature includes (1) \mtranse~\cite{DBLP:conf/ijcai/ChenTYZ17}, which proposes to utilize \textsf{TransE} for EA; (2) \iptranse~\cite{DBLP:conf/ijcai/ZhuXLS17}, which uses an iterative training process to improve the alignment results; (3) \bootea~\cite{DBLP:conf/ijcai/SunHZQ18}, which devises an alignment-oriented KG embedding framework and a bootstrapping strategy; (4) \rsn~\cite{DBLP:conf/icml/GuoSH19}, which integrates recurrent neural networks with residual learning; 
(5) \mc~\cite{DBLP:conf/acl/CaoLLLLC19}, which puts forward a multi-channel graph neural network to learn alignment-oriented KG embeddings; (6) \kecg~\cite{li2019semi}, which proposes to jointly learn knowledge embeddings that encode inner-graph relationships, and a cross-graph model that enhances entity embeddings with their neighbors' information; and (7) \te~\cite{DBLP:conf/semweb/SunHHCGQ19}, which presents a novel edge-centric embedding model that contextualizes relation representations in terms of specific head-tail entity pairs.

Methods incorporating other types of information include (1) \jape~\cite{DBLP:conf/semweb/SunHL17}, where attributes of entities are harnessed to refine the structural information; (2) \gcn~\cite{DBLP:conf/emnlp/WangLLZ18}, which generates entity embeddings and attribute embeddings to align entities in different KGs; (3) \gm~\cite{ACL19}, where a local subgraph of an entity is constructed to represent entity, and entity name information is harnessed for initializing the framework; (4) \mul~\cite{IJCAI19}, which offers a novel framework that unifies the views of entity names, relations and attributes at \emph{representation-level for mono-lingual} EA; (5) \rd~\cite{DBLP:conf/ijcai/WuLF0Y019}, which proposes a relation-aware dual-graph convolutional network to incorporate relation information via attentive interactions between KG and its dual relation counterpart; and (6) \hgcn~\cite{wu2019jointly}, where a learning framework is built to jointly learn entity and relation representations for EA.

\subsection{Results} \label{sec:results}
Table~\ref{tab:EA} summarizes the results.
Solutions in the first group merely harness structural information for aligning.
\bootea and \kecg achieve better performance than \mtranse and \iptranse due to the carefully designed alignment-oriented KG embedding framework and attention-based graph embedding model, respectively.
\rsn further advances the results by taking into account long-term relational dependencies between entities, which can capture more structural signals for alignment.
\te attains the best performance due to its edge-centric KG embedding and the bootstrapping strategy.
\mc fails to yield promising results as there are no aligned relations on $SRPRS$, where rule transferring cannot work, and the number of detected rules is rather limited.
Noteworthily, Hits@1 values on most datasets are below 50\%, showing the inefficiency of solely relying on KG structure, especially when long-tail entities account for the majority.

Regarding the second group, both \gcn and \jape exploit attribute information to complement structural signals. However, they fail to outperform the leading method in the first group, which can be attributed to the limited effect of attributive information.
The rest four methods all employ the generally available entity name information.
The significant performance gain compared with the first group validates the usefulness of this feature.
Among others, \our outperforms \gm, \rd and \hgcn by approximately 10\% on Hit@1 over all datasets, verifying that our framework can best exploit entity name information.
The underlying reason is that \gm, \rd and \hgcn fuse features on representation level, which potentially causes the loss of information as the resulting unified feature representation cannot keep the characteristics of original features.
\our, however, utilizes a co-attention network to determine feature weights and fuses features at outcome level, namely, feature-specific similarity scores.

\subsubsection*{Evaluation by degree}
To demonstrate that \our is effective at aligning long-tail entities, 
we show the results by degree in Table~\ref{tab:deg}.
Note that for \our, the degree refers to the initial degree distribution (as the completion process alters entity degrees).

From Table~\ref{tab:deg}, it reads that for entities with degree 1, the Hits@1 values of \our are two or even three times higher than those of \rsn, validating the effectiveness of \our for dealing with long-tail issue.
Although for popular entities, the performance also increases, the improvement over \rsn narrows greatly in comparison to the entities in tail.
Moreover, \our outperforms \rd across all degree groups in four datasets, despite that both of them utilize entity name information as an external signal for EA.

\subsubsection*{Comparison with \mul on dense datasets}
The results of \mul on $SRPRS$ are not provided since it can merely cope with mono-lingual dataset and demands semantics of relations to be known in advance. 
To better appreciate \our, we supply the experimental results of \our on the \emph{dense} datasets where \mul was evaluated.
Concretely, the dense datasets, DBP-WD-100K and DBP-YG-100K, are similar to DBP-WD and DBP-YG, but some with larger scale (100K entities on each side) and higher density~\cite{DBLP:conf/ijcai/SunHZQ18}.

\begin{table}[t]
	\centering \small
	\caption{Experiment results on dense datasets}
	\begin{tabular}{lcccccc}
		\toprule
		\multirow{2}[4]{*}{Methods} & \multicolumn{3}{c}{DBP-WD-100K} & \multicolumn{3}{c}{DBP-YG-100K} \\
		\cmidrule{2-7}          & Hits@1 & Hits@10 & MRR   & Hits@1 & Hits@10 & MRR \\
		\midrule
		\mul & 91.9 & 96.3 & 0.94  & 88.0 & 95.3 & 0.91 \\
		\our & \textbf{97.4} & \textbf{99.6} & \textbf{0.98} & \textbf{94.3} & \textbf{98.6} & \textbf{0.96} \\
		\bottomrule
	\end{tabular}%
	\label{tab:addlabel}%
\end{table}%

On dense datasets, \our yields better results, with Hits values over 90\% and MRR over 0.95 (shown in Table~\ref{tab:addlabel}), 
demonstrating that \our better exploits name information, which can be attributed to our degree-aware feature fusion module and the choice of calculating scores within each view first, instead of learning a combined representation that might cause information loss.

\begin{table}[h]
	\centering \small
	\caption{Experiment results of ablation}
	\begin{tabular}{lccc}
		\toprule
		\multirow{2}[4]{*}{Methods} & \multicolumn{3}{c}{EN-FR} \\
		\cmidrule{2-4}          & Hits@1 & Hits@10 & MRR \\
		\midrule
		\our   & \textbf{75.8}  & \textbf{89.9}  & \textbf{0.81} \\
		\our w/o IKGC  & 72.1  & 85.4  & 0.77 \\
		\our w/o KGC   & 73.9  & 88.6  & 0.79 \\
		\our w/o ATT   & 73.1  & 88.5  & 0.79 \\
		\our w/o CPM   & 75.3  & 89.7  & 0.80 \\
		\bottomrule
	\end{tabular}%
	\label{tab:feature}%
\end{table}%

\subsection{Ablation study}\label{abla}
We report an ablation study on EN-FR dataset in Table~\ref{tab:feature}.

\begin{figure} [t]
	\centering
	\includegraphics[width=0.8\linewidth]{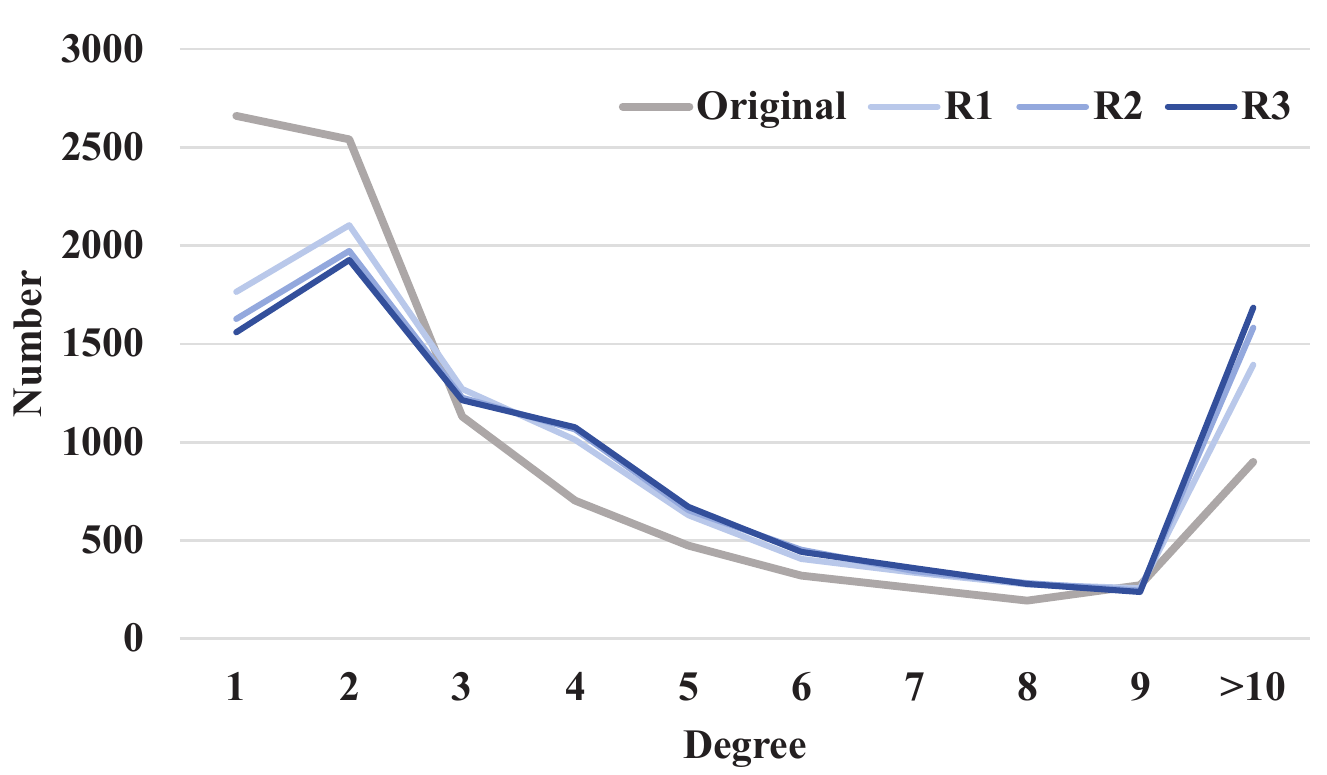}
	\caption{Distribution of entity degree in EN-FR}\label{fig:dis}
\end{figure}

\subsubsection*{Iterative KG completion}
By removing the entire module, EA performance drops by 3.7\% on Hits@1 (\our vs. \our w/o IKGC).
If we merely eliminate the KG completion module while keep the iterative process (similar to~\cite{DBLP:conf/ijcai/ZhuXLS17}), Hits@1 also declines by 1.9\% (\our vs. \our w/o KGC), validating the significance of KG completion.
We further show the dynamic change of degree distribution after each round, i.e., Original, R1, R2, R3, in Figure~\ref{fig:dis}, 
which indicates that the embedded KG completion enhances KG coverage and reduces the number of long-tail entities.

\begin{figure} [b]
	\centering
	\includegraphics[width=\linewidth]{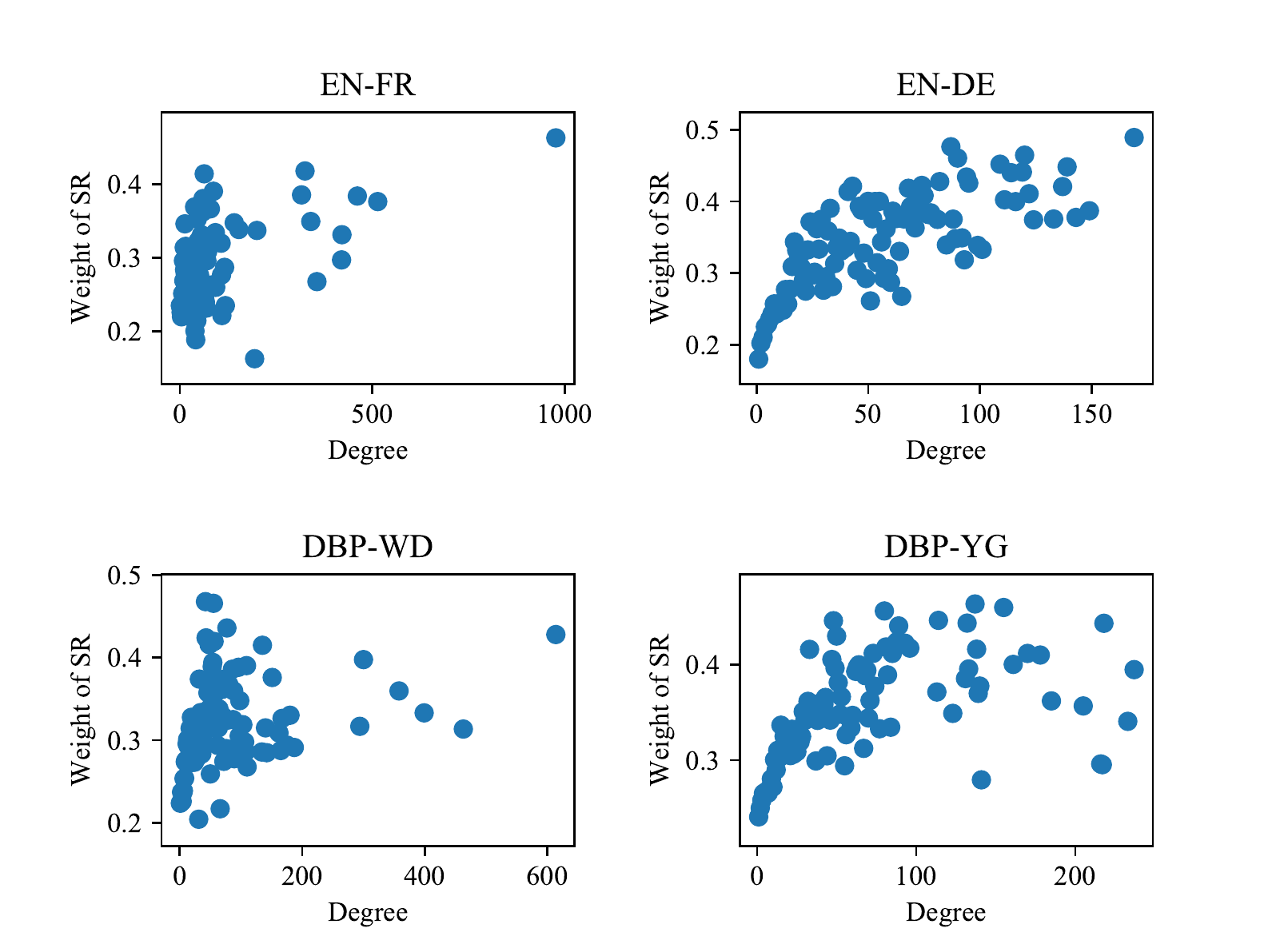}
	\caption{Weight distribution of structural representation}\label{fig:weight}
\end{figure}

\begin{figure*}[h]
	\centering
	\subfigure[Number of pairs selected]{
		\centering
		\includegraphics[width=5cm]{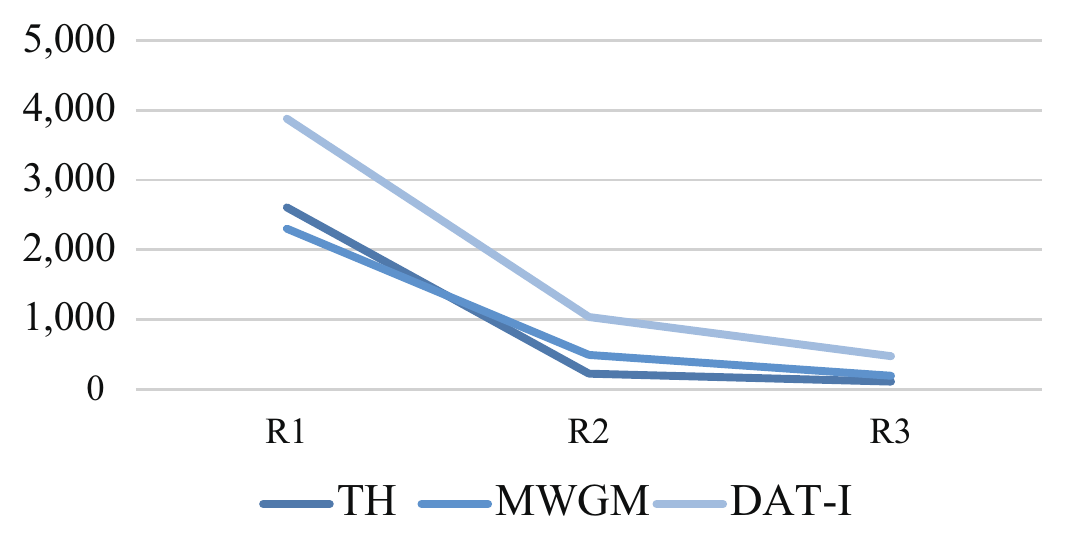}
	}%
	\goodgap \goodgap \goodgap
	\subfigure[Accuracy of pairs selected]{
		\centering
		\includegraphics[width=5cm]{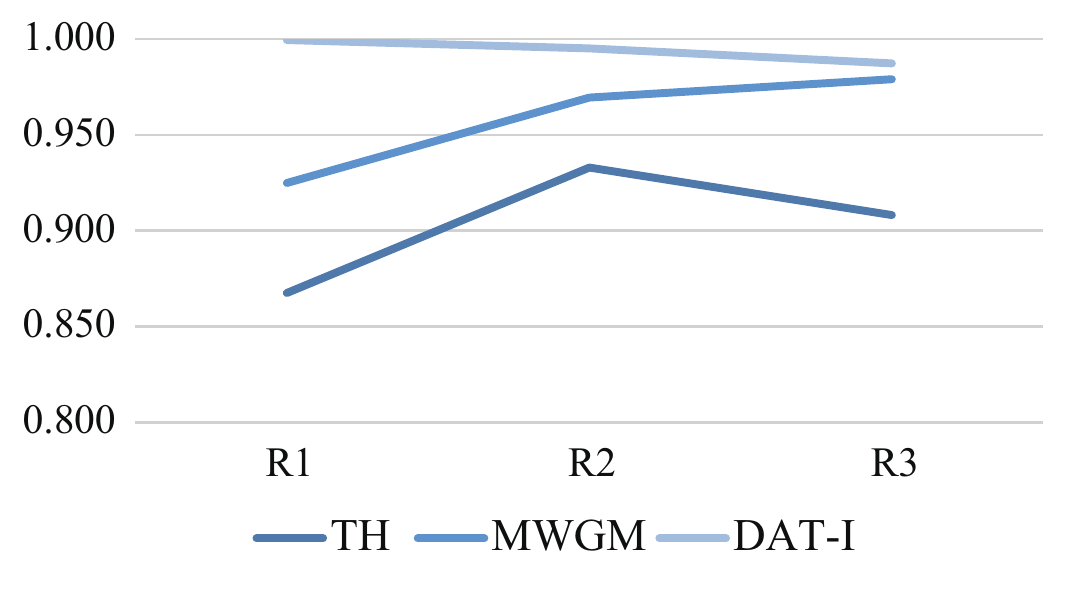}
	}%
	\goodgap \goodgap \goodgap
	\subfigure[Running time consumption (s)]{
		\centering
		\includegraphics[width=5cm]{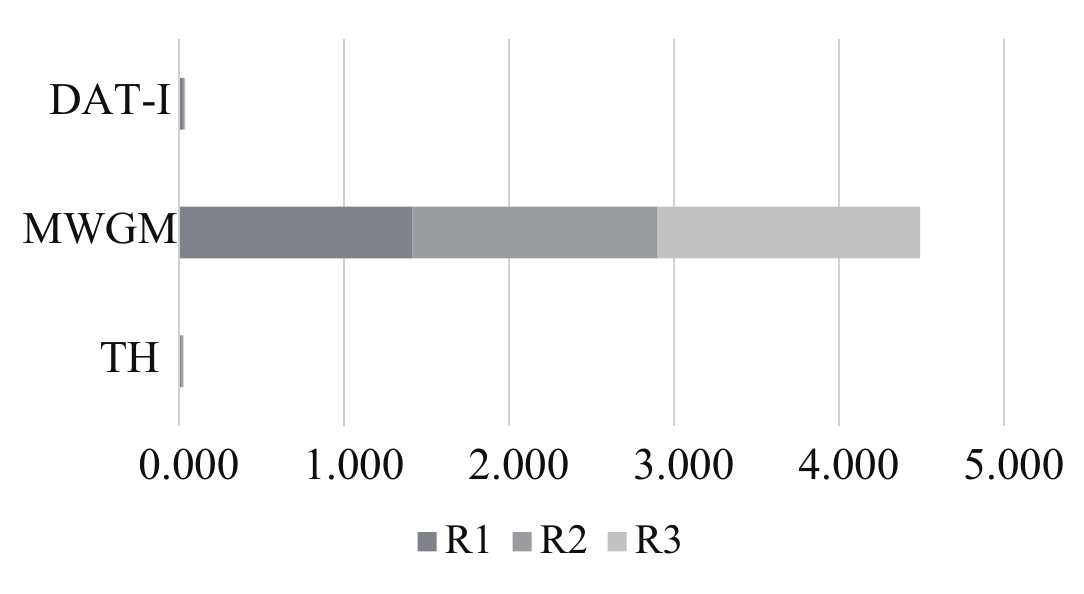}
	}
	
	\centering
	\caption{Comparison results of iterative training strategies}\label{fig:ite}
\end{figure*}

\subsubsection*{Degree-aware co-attention feature fusion}
Table~\ref{tab:feature} shows that replacing {\itshape Degree-Aware Fusion Module} with fixed equal weights results in a 2.7\% drop on Hits@1 (\our vs. \our w/o ATT).
This validates that dynamically adjusting the weights of features according to the degree leads to better feature integration and hence more accurate alignment results.
We further display the weight of structural representation generated by our degree-aware fusion model across different degrees (in 1st round) in Figure~\ref{fig:weight}.
It reveals that, by and large, the significance of structure information increases as the degree of entities rises, which is in accordance with our anticipation.

\subsubsection*{Concatenated power mean word embeddings}
We compare concatenated power mean word embeddings and averaged word embeddings against aligning entities (\our vs. \our w/o CPM). The results suggest that stacking different power mean embeddings does capture more features for alignment.

\subsection{Error analysis}\label{error}
We perform error analysis on EN-FR dataset to demonstrate the contribution of each module, as well as cases where \our fails.
On EN-FR, solely using structural information leads to 65.5\% error rate on the Hits@1. 
67.0\% of the entities are long-tail (i.e., with degree $\leq$ 3), among which 65.1\% are wrongly aligned. 
Introducing entity name information and dynamically fusing it with structural information reduces the overall Hits@1 error rate to 27.9\%.
For long-tail entities, the error rate is also reduced to 33.2\%. 
On top of it, employing iterative KG completion to replenish structure and propagate the signals can further reduce the overall Hits@1 error rate to 24.2\%.
The percentage of long-tail entities is also reduced to 49.7\%, of which merely 8.3\% are wrongly aligned.
It implies that long-tail entities account for the most errors initially, and then applying the proposed techniques reduces not only the error rate but also the portion of (contributing) long-tail entities. 

As for the very difficult cases that \our cannot resolve, we give following analysis with the focus on entity name information.
Among the false cases (24.2\% on EN-FR), 41\% do not have an proper entity name embedding as all words in the name are OOVs, and 31\% suffer from partial OOVs.
Moreover, 15\% could have achieved correct results by solely using name information but get misled by structural signals,
and 13\% fail to align due to the insufficiency of the entity name representation approach or the fact that the entities of the identical name denote different physical ones.

\subsection{Further experiment}\label{feature-analysis}
We further justify the effectiveness of our iterative training strategy by conducting the following experiments.

Our iterative process differs from existing methods in not only the embedded KG completion process but also the selection of confident pairs.
To demonstrate the merit, we exclude the KG completion module from \our resulting \textsf{DAT-I}, to compare with the selection methods of ~\cite{DBLP:conf/ijcai/ZhuXLS17,DBLP:conf/ijcai/SunHZQ18}.
In~\cite{DBLP:conf/ijcai/ZhuXLS17}, for each \emph{non-aligned} source entity, it finds the most similar \emph{non-aligned} target entity, and if the similarity between the two entities is above a given threshold, it is confident to consider these two as a pair. We refer to this as threshold-based method (\textsf{TH}).
In~\cite{DBLP:conf/ijcai/SunHZQ18}, for each source entity, it computes alignment likelihood to every target entity, and only those with likelihood above a given threshold are involved in a maximum likelihood matching processing under 1-to-1 mapping constraint, which generates a solution containing confident EA pairs. We refer to this as maximum weight graph matching (\textsf{MWGM}).
We re-implement the methods under our framework and tune the parameters according to their original papers. 
To measure the effectiveness of different iterative training methods, we adopt the number of selected confident EA pairs, the accuracy of these EA pairs and the running time of each round as the main metrics.

For fair comparison, we report the results of first three rounds in Figure~\ref{fig:ite}.
It is observed that, \textsf{DAT-I} outperforms the other two in terms of the number and the quality of selected pairs in comparatively less time. 
\textsf{MWGM} requires much more time as it needs to solve a global optimization problem, whereas compared with \textsf{TH}, it performs better in terms of the accuracy of selected pairs.

\section{Conclusion}\label{conclude}
In this work, we offer a revisited framework \our for entity alignment with emphasis on long-tail entities.
Observing the limit of sole reliance on structural information, we propose to introduce entity name information in the form of concatenated power mean embedding in pre-alignment phase.
In alignment, to consolidate different signals, we devise a co-attention feature fusion network that dynamically adjusts the weights of different features under the guide of degree.
During post-alignment, we complete KG with knowledge from the other side using confident EA results as anchors in an iterative fashion, which amplifies the structural information and boosts the performance. 
\our is evaluated on both cross-lingual and mono-lingual EA benchmarks and achieves superior results.

\begin{acks}
This work was partially supported by NSFC under grants Nos. 61872446, 61902417, and 71971212, and NSF of Hunan province under grant No. 2019JJ20024, and Postgraduate Scientific Research Innovation Project of Hunan Province (CX20190033, CX20190032). 
Wei Wang was partially funded by ARC DPs 170103710 and 180103411, and D2DCRC DC25002 and DC25003.
\end{acks}

\bibliographystyle{ACM-Reference-Format}
\bibliography{sample-base}


\end{document}